\documentclass{llncs}
\usepackage{makeidx}  
\usepackage{times}
\usepackage{helvet}
\usepackage{courier}
\usepackage{graphicx}
\usepackage{bm}
\usepackage{amsfonts}
\usepackage{amsmath}
\usepackage{extarrows}
\usepackage{bbm}
\usepackage{blindtext, subfig}
\usepackage{dblfloatfix}
\newtheorem{thm}{Theorem}
\newtheorem{prf}{Proof}

\usepackage{tikz}
\usetikzlibrary{arrows}
\begin{document}

\mainmatter              
\title{A Generalized Model for Multidimensional Intransitivity}
%

\author{Jiuding Duan \and Jiyi Li
\and Yukino Baba \and Hisashi Kashima}
\institute{Department of Intelligence Science and Technology\\ Kyoto University, Kyoto, 606-8501, Japan,\\
\email{dj@ml.ist.i.kyoto-u.ac.jp, \{jyli,baba,kashima\}@i.kyoto-u.ac.jp}
}

%
%


\maketitle              

\begin{abstract}
Intransitivity is a critical issue in pairwise preference modeling. It refers to the intransitive pairwise preferences between a group of players or objects that potentially form a cyclic preference chain, and has been long discussed in social choice theory in the context of the dominance relationship. However, such multifaceted intransitivity between players and the corresponding player representations in high dimension are difficult to capture. In this paper, we propose a probabilistic model that joint learns the $d$-dimensional representation ($d > 1$) for each player and a dataset-specific metric space that systematically captures the distance metric in $\mathbb{R}^d$ over the embedding space. Interestingly, by imposing additional constraints in the metric space, our proposed model degenerates to former models used in intransitive representation learning. Moreover, we present an extensive quantitative investigation of the wide existence of intransitive relationships between objects in various real-world benchmark datasets. To the best of our knowledge, this investigation is the first of this type. The predictive performance of our proposed method on various real-world datasets, including social choice, election, and online game datasets, shows that our proposed method outperforms several competing methods in terms of prediction accuracy.
\end{abstract}
\let\thefootnote\relax\footnotetext{This paper is published at the 21st Pacific-Asia Conference on Knowledge Discovery and Data Mining (PAKDD), Jeju, South Korea, 2017} 

\section{Introduction}
The {\it transitivity} of pairwise comparison and matchup between individual objects is a fundamental principle in both social choice theory \cite{tideman2006collective,law1927} and preference data modeling \cite{regenwetter2011transitivity}.

In pairwise comparison, two participants in a single round are evaluated by a third-party judge or an objective rule that judges the discriminative \textsl{win/lose} result for each player. Examples of applications of such a comparison include recommender systems \cite{jamali2011transitivity}, social choice systems \cite{herlocker1999algorithmic,kamishima2003nantonac,tideman2006collective}, and so on. In pairwise matchup, two participants are each other's competitive opponents, and therefore the discriminative win/lose result is a reflection of their strength in the game. Examples of such matchup applications are sports tournaments \cite{cattelan2013dynamic} and online games \cite{chen2016modeling}. In both cases, the hidden winning ability of each individual object can be quantitatively profiled by parametric probabilistic models \cite{law1927,bradley1952rank}.

\begin{figure}
\begin{minipage}[t]{.3\textwidth}
\vspace{0pt}
  \includegraphics[width=\linewidth]{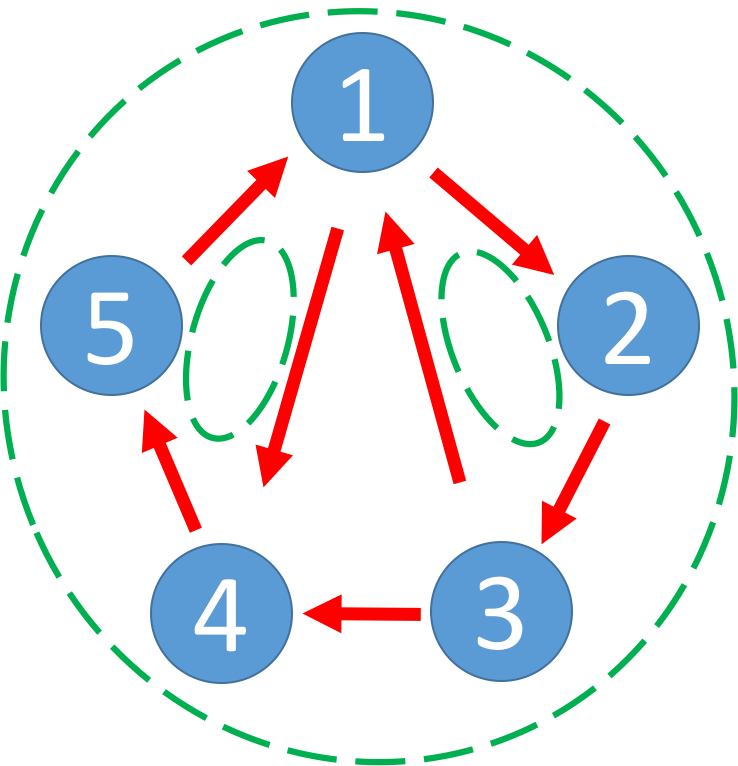}
  \caption{Directed asymmetric graph illustration of the observed game in Table \ref{fig:nuisance}}
  \label{fig:nuisance}
\end{minipage}\hfill
\begin{minipage}[t]{.7\textwidth}
\vspace{0pt}
\label{tab:nuisance}
\centering
\captionof{table}{Toy model demonstrating the subtle deterioration in terms of test accuracy}
\begin{tabular}{c|c|c|c|c|c|c} \hline
Winner ID & Loser ID & \#wins & \#loses & GT & $\text{pred}_{trans}$ & $\text{pred}_{intrans}$\\ \hline
1 & 2 & 10 & 5 & \checkmark & \checkmark & \checkmark\\ \hline
1 & 3 & 1 & 2 & \checkmark & x & \checkmark\\ \hline
1 & 4 & 10 & 5 & \checkmark & \checkmark & \checkmark\\ \hline
1 & 5 & 1 & 2 & \checkmark & x & \checkmark\\ \hline
2 & 3 & 10 & 5 & \checkmark & \checkmark & \checkmark\\ \hline
3 & 4 & 10 & 5 & \checkmark & \checkmark & \checkmark\\ \hline
3 & 5 & 10 & 5 & \checkmark & \checkmark & \checkmark\\ \hline
4 & 5 & 10 & 5 & \checkmark & \checkmark & \checkmark\\ \hline
\hline
\multicolumn{5}{c|}{Test Accuracy} & 0.6458 & 0.6667 
\end{tabular}
\end{minipage}
\end{figure}

However, in addition to the thorough theoretical justifications of these parametric probabilistic models that assume certain levels of transitivity, the existence of {\it intransitivity}, which overrides the transitivity of preference in the real world, has been argued in ecometrics, behavior economics, and social choice theory for decades \cite{may1954intransitivity,tideman2006collective}. Intransitivity refers to the property of binary relations (i.e., win/loss or like/dislike) that are not transitive. For instance, in a rock-paper-scissors game, the pairwise matchup result is judged by three rules: \{$o_{paper} \succ o_{rock}$, $o_{rock} \succ o_{scissors}$, and $o_{scissors} \succ o_{paper}$ \}. A transitive model results in a transitive dominance $o_{paper} \succ o_{scissors}$, that violates the third rule $o_{scissors} \succ o_{paper}$. In other words, the binary relations in the rock-paper-scissors game are not transitive. Such intransitivity in the real world exists in the form of cyclic dominance that implies the non-existence of a local dominant winner in the local preference loop. In many applications, the presence of a nested local intransitive preference loop results in systematically intransitive comparisons and matchups, and therefore predictive modeling is challenging. Intuitively, this situation occurs when objects have multiple features or views of judgment and each of these views dominates a corresponding pairwise comparison. The underestimation of such cyclic dominance is subtle in the numerical testing scores in terms of prediction accuracy, but critical for the cost-sensitive decision making based on the prediction results, as illustrated in the toy model in Figure \ref{fig:nuisance} and Table \ref{fig:nuisance}.

Figure \ref{fig:nuisance} shows a directed asymmetric graph (DAG) to illustrate the toy game records in Table \ref{fig:nuisance}; the numbered nodes represent the corresponding player, the arrows demonstrate the dominant relationship between players, and the three dotted circles demonstrate the existing cyclic intransitive dominance relationships in the observed game records. In Table \ref{fig:nuisance}, the last two columns are exemplar predictions derived from transitive and intransitive models. The prediction of a transitive model $\text{pred}_{trans}$ cannot fully capture the intrinsic intransitivity in the dataset, leading to a deterioration in terms of predictive performance, whereas the prediction made by an intransitivity-compatible model $\text{pred}_{intrans}$ is able to accurately capture all the deterministic matchups. The mis-prediction of two out of the eight relationships results in only a subtle deterioration of the average test accuracy by 0.0208. Moreover, a growth in the number of observed records leads to a further difficulty in the evaluation of the unveiled intransitivity. In this toy model, the local intransitive sets \{1,2,3\} and \{1,4,5\} are nested in a global intransitive set \{1,2,3,4,5\}. Such $c$ locally nested structures in a dataset with a large number of players $n$ and active dominance $e$ lead to an exponentially growing number of intransitive cycles. The most efficient algorithm for searching all such cycles yields a time complexity bounded by $O((n+e)(c+1))$ \cite{johnson1975finding}, which is intractable for stochastically observed dense matchups with large numbers of participants. Thus, the approach of modeling the multidimensional intransitive embedding by ensemble learning of all the possible views is blocked. A detailed quantitative exploration of the cyclic intransitivity in a variety of real-world datasets is presented in later sections.

The challenge preseneted by intranstivity motivated the alternative approach of learning the intransitivity-compatible multidimensional embedding from the parametric probabilistic models for pairwise comparisons. Without loss of generality, we attribute both pairwise comparison and matchup to the single notion of \textsl{matchup} and denote the individuals in the matchup as \textsl{players} in the following context, and discuss only the non-tie case for simplicity.


Existing work in this line of research includes studies on the seminal Bradley-Terry (BT) pairwise comparison model \cite{bradley1952rank} and its extensions and applications in various real-world data science applications, e.g., matchup prediction \cite{cattelan2013dynamic}, social choices \cite{herlocker1999algorithmic,kamishima2003nantonac,tideman2006collective}, and so on. In the BT model, the strength of the players is parameterized as a single scalar value, by which the matchups between players always remain transitive. Other attempts to meet the challenge include extending the scalar into a 2-dimensional vector representation through a non-linear logistic model \cite{2dextensionbt2005}, and the more recently proposed Blade-Chest (BC) model with a multidimensional embedding scheme that imitates the offense and defense ability of a player in two independent multidimensional spaces \cite{chen2016modeling,chen2016incontext}. However, the BC model, which was extended directly from the seminal BT model, is limited in its expressiveness of intransitivity by the arbitrary separation of the two representation metric spaces and an unexpected numerical conjugation drawback.

In this paper, we address the problem of predictive modeling of the intransitive relationships in real-world datasets by learning the multidimensional intransitivity representation for each player, i.e., items in a recommender system, tennis players in a tennis tournament, game players in online game platforms, or candidates in a political election. We focus on joint learning of the $d$-dimensional representation ($d > 1$) for each player and a dataset-specific metric space that systematically captures the distance metric in $\mathbb{R}^d$ over the embedding space. The joint modeling of the multidimensional embedding representation and the metric space is achieved by involving two types of covariate matrices, one to capture the interactive battling result between two players on the metric space, and a second to capture the intrinsic strength of each player. Through an analysis of the symmetry and expressiveness of our proposed embedding formulation, we further argue that the constrained optimization problem that is induced by our proposed multidimensional embedding formulation can be indentically transformed into an unconstrained form, thus allowing a generic numerical solution of the proposed model by using a stochastic gradient descent method \cite{bottou2010large}. Finally, we evaluate the effectiveness of our proposed method on a variety of real-world datasets, and demonstrate its superiority over other competitive methods in terms of predictive performance.

Our contributions are as follows:
\begin{itemize}
	\item An extensive investigation of the wide existence of intransitive relationships between objects in many prevalent real-world benchmark datasets. This investigation required that special attention be paid to intransitive relationships. To the best of our knowledge, this is the first quantitative exploration of the existing intransitive relationships in these prevalent benchmark datasets, and even the first in the data mining research community.
	\item The proposal of a generalized embedding formulation for learning the intransitivity-compatible representation from pairwise matchup data, and an efficient solution to the induced optimization problem, together with a systematic characterization of the model, bridging the proposed generalized model and the former multidimensional representation learning methods. 
	\item An empirical evaluation of the proposed method on various real-world datasets, which demonstrates the superior performance of the proposed method in terms of prediction accuracy.
\end{itemize}

The rest of the paper is organized as follows. Section 2 presents the related work on modeling intransitive relationships from pairwise comparison data. Section 3 defines the representation learning problem and presents our generalized formulation of the multidimensional embedding. In Section 4 we describe our investigation of the existence of intransitivity in the real-world datasets and present the experimental results for both synthetic and real-world datasets. Section 5 conlcudes our paper.

\section{Related Work}
Existing work on parametric models for pairwise matchups data, which originate from seminal work performed decades ago and include the Thurstone model \cite{law1927} and the Bradley-Terry-Luce model \cite{bradley1952rank}, were surveyed extensively. The BT model \cite{bradley1952rank} is based on maximum likelihood estimation and was further generalized to multiparty matchups \cite{generalizedbt2004} and adapted to comparisons involving a tie \cite{davidson1970extending}. The first BT model generalized to multi-dimensional representation was limited to the 2-dimensional case with a non-linear logistic function, inspired by classical multidimensional scaling \cite{2dextensionbt2005}. In real-world matchups, the ranking of the players' ability is an issue that is closely related to our parametric modeling for pairwise matchup data. Especially in sports tournaments \cite{football1993,cao2007learning,groupcomparison2008} and online games \cite{trueskill2006}, the Elo ratings system \cite{elo1978rating} and the TrueSkill ratings system \cite{trueskill2006,trueskilltime2007} are noteworthy. In addition, instead of modeling the matchups between individual players, some methods concentrate on group matchup \cite{football1993,groupcomparison2008}, rating individual players from the group matchup records \cite{ANNIndividualGroup2008}, or alternatively model the belief of each collected record \cite{crowdbt}. These methods are different from ours in that they were all developed according to the principles of transitivity.

In the context of modeling intransitivity, by extending the BT model, a 2-dimensional vector can be employed as the ability of players in matchups \cite{2dextensionbt2005}, with no verification of the modeling of the intransitive relationships on large datasets. The state-of-the-art model for intransitive modeling is the BC model \cite{chen2016modeling}, which imitates the offense and defense characteristics of a player and learns the corresponding multidimensional representations from matchup records. The BC model was then further extended to contexture-aware settings \cite{chen2016incontext} with an improvement in the performance. 



\section{Proposed model}
Assume a given set of candidate players $\mathbf{P}$ with $|\mathbf{P}| = M$. The dataset $\mathbf{D}$ contains $N$ pairwise matchup records $x_i(a_i,b_i) \in \{0,1\}$, $i = [1\text{:}N]$, where the players $a_i$ and $b_i$ $\in \mathbf{P}$. An ordinal matchup record $o_a \succ o_b$ is the matchup record between player $a$ and player $b$, meaning $a$ beats $b$, and $o_a \prec o_b$ , vice versa. The observed record $x(a,b)$ can be represented in a 4-tuple: either $x(a,b) = (a,b,1,0)$ meaning $o_a \succ o_b$ or $x(a,b)=(a,b,0,1)$ meaning $o_a \prec o_b$. The identical deterministic events can be aggregated, resulting in a collapsed dataset $\mathbf{D}^{collapse}$. The data entry $x_{aggregate}(a,b) \in \mathbf{D}^{collapse}$ is given by 4-tuples in $x_{aggregate}(a,b) = (a,b,n_a,n_b)$, where $n_a$ is the total count of observed event $o_a \succ o_b$, and $n_b$ of $o_a \prec o_b$, accordingly. 

The goal is to predict the result of matchups by learning the interpretable multidimensional representation of the players, that reflects their ability in multiple views.

\subsection{Bradley-Terry Model and Blade-Chest Model}
In the BT model, each player $p \in \mathbf{P}$ is parameterized by a scalar $\gamma_p \in \mathbb{R}$ as the indicator of his/her ability to win. Following the probability axiom, the probability of the event is modeled as
\begin{eqnarray}
Pr(o_a \succ o_b) &=& \frac{\exp(\gamma_a)}{\exp(\gamma_a) + \exp(\gamma_b)}\\
&=& \frac{1}{1+\exp(-M_{ab})}
\end{eqnarray}
where $M_{ab} = \gamma_a - \gamma_b$ is the symmetric {\it matchup function} for player $a$ and player $b$, with property
\begin{eqnarray}
\label{eq: sym}
M_{ab} = - M_{ba}
\end{eqnarray}
and
$$Pr(o_a \prec o_b) = 1 - Pr(o_a \succ o_b) $$

The scalar-valued ability indicator of players $\gamma_p$ is not intransitivity-aware and this has been shown in various datasets \cite{may1954intransitivity,chen2016modeling}. The parameter estimation of the BT model can be conducted by applying an EM algorithm for maximum likelihood or more generalized techniques \cite{generalizedbt2004}. Note that the matchup function $M_{ab}, a,b \in \mathbf{P}$ is the learning oracle that accesses the latent metric of players’ ability, and therefore, it can be further extended to a multidimensional setting, named the BC model \cite{chen2016modeling}. Intransitivity is then embraced by the BC model, where blade and chest vectors imitate the offense and defense, respectively. 

Formally, in the BC model, the ability of player $p \in \mathbf{P}$ is parameterized by $\mathbf{a}_{blade}$ and $\mathbf{a}_{chest} \in \mathbb{R}^d$ and the corresponding matchup function is formulated by
\begin{itemize} 
\item the Blade-Chest-Inner (BCI) embedding $M^{BCI}(a,b)$
\begin{eqnarray}
\label{eq: bci}
M^{BCI}(a,b) = \mathbf{a}_{blade}^T\cdot\mathbf{b}_{chest}-\mathbf{b}_{blade}^T\cdot\mathbf{a}_{chest}
\end{eqnarray}
\item the Blade-Chest-Distance (BCD) embedding $M^{BCD}(a,b)$
$$M^{BCD}(a,b) = \left\|\mathbf{b}_{blade}-\mathbf{a}_{chest}\right\|_{2}^{2}-\left\|\mathbf{a}_{blade}-\mathbf{b}_{chest}\right\|_{2}^{2}$$
\end{itemize}

These formulations of the matchup function naturally ensure the symmetry property denoted in Condition (\ref{eq: sym}), and therefore are compatible with the scalar-valued representation of the players' strength in the BT model. The connection between these two formulations can also be evidenced under a mild condition \cite{chen2016modeling}. Assembled by this multidimensional formulation, the BC model is state-of-the-art in both predictive modeling and representation learning for the players' intransitivity.

\subsection{Generalized Intransitivity Model}
We propose a generic formulation of the matchup function that jointly captures a $d$-dimensional representation ($d > 1$) for each player and a dataset-specific distance metric for the learned representation in $\mathbb{R}^d$ over the embedded dimensions. Let us assume we have a $d$-dimensional representation $\mathbf{a} \in \mathbb{R}^d$ for player $a \in \mathbf{P}$; then, we formulate the generalized intransitivity embedding $M^{G}(a,b)$ as,
\begin{eqnarray}
\label{eq:gen_intrans}
M^{G}(a,b) &=& \mathbf{a}^{T}\Sigma\mathbf{b} + \mathbf{a}^{T}\Gamma\mathbf{a} - \mathbf{b}^{T}\Gamma\mathbf{b}
\end{eqnarray}
where $\mathbf{a}$ and $\mathbf{b}$ are the $d$-dimensional representation for player $a$ and player $b$, respectively, and $\Sigma,\Gamma \in \mathbb{R}^{d\times d}$ are the {\it transitive matrices}. The model parameters we attempt to learn are $\theta^{G} := \{\mathbf{a},\mathbf{b},\Sigma,\Gamma\}$. In the proposed formulation, the first term $\mathbf{a}^{T}\Sigma\mathbf{b}$ reflects the interaction between players, and the latter term $\mathbf{a}^{T}\Gamma\mathbf{a} - \mathbf{b}^{T}\Gamma\mathbf{b}$ reflects the intrinsic strength of each individual. The embedding is proposed to model the pairwise preference, in which two properties should be preserved, i.e., preference symmetry and expressiveness.

\subsection{Properties}
We characterize the detailed properties of the proposed formulation in terms of symmetry and expressiveness in comparison with the BC model, and show that the BC model is a specialized formulation in a family of our generalized formulation.

\subsubsection{Symmetry}
Since we discuss the matchup result between two players, the symmetry must be preserved \cite{law1927}. This is different from other problems, such as link prediction in social networks, where the directed preference between items is naturally asymmetric \cite{Ou:2016:ATP:2939672.2939751}.

Obviously, the two numerical computations of the first term $\mathbf{a}^{T}\Sigma\mathbf{b}$ and the latter term $\mathbf{a}^{T}\Gamma\mathbf{a} - \mathbf{b}^{T}\Gamma\mathbf{b}$ are independent given randomized $d$-dimensional embeddings $\mathbf{a}$ and $\mathbf{b}$. Without intuition of the specific design of $\mathbf{a}$ and $\mathbf{b}$, a.k.a. random initialization, the sufficient condition to preserve the symmetry of the first term is
\begin{eqnarray}
\label{eqn:transposeEqual}
\Sigma = - \Sigma ^T
\end{eqnarray}
which is difficult to regularize given the gradient $\nabla_{\Sigma}M^G(a,b)$:
$$\nabla_{\Sigma}M^G(a,b) = \mathbf{a}\mathbf{b}^{T}$$


However, if we introduce it as a constraint in the optimization, the induced constrained optimization problem is difficult to solve. Alternatively, we devise an efficient solution which transforms the constrained optimization problem into an unconstrained optimization by reparameterizing $\Sigma$ with $\Sigma'$ by
\begin{eqnarray}
\label{re-parameterize}
\Sigma = \Sigma' - {\Sigma'}^T
\end{eqnarray}
where $\Sigma'$ is a free matrix having the same shape as $\Sigma$. To this end, it is trivial to show that the symmetry of $\mathbf{a}^{T}\Sigma\mathbf{b}$ is preserved. Together with the fact that the symmetry of the self-regulation term $\mathbf{a}^{T}\Gamma\mathbf{a} - \mathbf{b}^{T}\Gamma\mathbf{b}$ in $M^{G}$ holds constantly, we conclude that the symmetry of the proposed matchup function formulation is guaranteed.

\subsubsection{Expressiveness}
We further characterize the superior expressiveness of our proposed intransitive representation learning technique. Interestingly, we show that the BC model is a specialized formulation within a family of our proposed formulation.

Suppose that we have blade and chest vectors for player $a$, $\mathbf{a}_{blade}$ and $\mathbf{a}_{chest}$ $\in \mathbb{R}^{d'}$, where $d' = 3$; then, we integrate them into a generalized vector $\mathbf{a}_{general}$ $\in \mathbb{R}^{2d'}$ defined by
\begin{eqnarray}
\label{eq:gen_vec}
&
\mathbf{a}_{general} = \begin{bmatrix}
\mathbf{a}_{blade} \\
\mathbf{a}_{chest} \\
\end{bmatrix}
= \begin{bmatrix}
blade_1\\
blade_2\\
blade_3\\
chest_1\\
chest_2\\
chest_3
\end{bmatrix}
\end{eqnarray}

This metaphorical definition is derived from the BC model, and therefore the $2d'$-dimensional generalized $\mathbf{a}_{general}$ has two distinct subspaces $\mathbf{a}_{blade}$ and $\mathbf{a}_{chest}$, which explicitly indicate the physical strength and weakness of player $a$, respectively.

\begin{thm}[Expressiveness]{Given the proposed matchup formulation in $2d'$-dimensional space, the proposed model degenerates to a BCI model in $d'$-dimensional space, under mild condition
\begin{equation}
\label{eq: norm_condition}
\left\|\mathbf{a}\right\|_{2}^{2} = \left\| \mathbf{b}\right\|_{2}^{2}
\end{equation}
$$\left\| \Gamma \right\|_{F} \rightarrow 0$$
and,
$$\Sigma = \begin{bmatrix}0& I_{d'\times d'} \\ -I_{d'\times d'}&0\end{bmatrix}$$
}
\end{thm}

\begin{prf}
{On the one hand, by the identified sufficient Condition (\ref{eqn:transposeEqual}) for the symmetry of $\mathbf{a}^{T}\Sigma\mathbf{b}$, given $I_{d'\times d'}$ as a $d'$-dimensional identity matrix, a fixed transitive matrix $\Sigma$ with
$$\Sigma = \begin{bmatrix}0& I_{d'\times d'} \\ -I_{d'\times d'}&0\end{bmatrix}$$
is a sufficient condition to preserve the symmetry of $\mathbf{a}^{T}\Sigma\mathbf{b}$, and results in
\begin{eqnarray}
\mathbf{a}^{T}\Sigma\mathbf{b} &=& \begin{bmatrix}
\mathbf{a}_{blade} \\
\mathbf{a}_{chest} \\
\end{bmatrix}^T 
\begin{bmatrix}0& I_{d'\times d'} \\ -I_{d'\times d'}&0\end{bmatrix}
\begin{bmatrix}
\mathbf{b}_{blade} \\
\mathbf{b}_{chest} \\
\end{bmatrix} \\
&=& \mathbf{a}_{blade}^T\cdot\mathbf{b}_{chest}-\mathbf{b}_{blade}^T\cdot\mathbf{a}_{chest}
\end{eqnarray}

On the other hand, given $\left\|\mathbf{a}\right\|_{2}^{2} = \left\| \mathbf{b}\right\|_{2}^{2} = c$, the inequality $\left\|\mathbf{a}^{T}\Gamma\mathbf{a} - \mathbf{b}^{T}\Gamma\mathbf{b}\right\| \le 2c\left\|\Gamma\right\| $ holds. Thus, $\mathbf{a}^{T}\Gamma\mathbf{a} - \mathbf{b}^{T}\Gamma\mathbf{b} \rightarrow 0$ holds by $\left\|\mathbf{a}\right\|_{2}^{2} = \left\| \mathbf{b}\right\|_{2}^{2} = c$ and $\left\| \Gamma \right\|_{F} \rightarrow 0$.

Therefore, the BCI model can be recovered by our proposed model. \qed
}
\end{prf}





Base on the fact that BCI formulation $M^{BCI}$ achieves better predictive performance than its variant $M^{BCD}$ in practice, and our proposed formulation $M^{G}$ degenerates into $M^{BCI}$ by imposing additional conditions, we argue that the proposed method is superior in terms of expressiveness over the BC model and the former models \cite{bradley1952rank,chen2016modeling}.

\subsection{Training}
Without loss of generality, given a set of players $\mathbf{P}$ and a collapsed training dataset $\mathbf{D}^{collapse}$ with pairwise matchup between players in 4-tuple $(a,b,n_a,n_b)$, as exemplified previously, our goal is to estimate the intransitivity parameters $\theta^G := \{\mathbf{a},\mathbf{b},\Sigma,\Gamma\}$ so that the predictive model can better predict unseen matchups. Following Equation (\ref{re-parameterize}), we reparameterize the transitive matrix $\Sigma$ as $\Sigma^\prime$ and optimize  $\theta^{G^\prime} := \{\mathbf{a},\mathbf{b},\Sigma',\Gamma\}$ instead. In line with the BT model, we train the model by maximum likelihood. The overall likelihood is given by
$$L(D|\theta^{G^\prime}) = \prod_{(a,b,n_{a},n_{b})\in\mathbf{D^{collapse}}} Pr(o_a \succ o_b )^{n_{a}}\cdot Pr(o_a \prec o_b )^{n_{b}}$$
where $Pr(o_a \succ o_b )$ is the probability of the event $o_a \succ o_b$. 

We take the log-likelihood and optimize it with a stochastic gradient descent method \cite{bottou2010large}, and randomly sample one 4-tuple from $\mathbf{D}^{collapse}$ in each epoch, and then update the model parameters $\theta^{G^\prime}$ w.r.t. the corresponding sample, until convergence.

\subsubsection{Regularization}
We choose the regularization terms as follows:
$$R_1(D|\theta^{G^\prime}) = \sum_{a\in\mathbf{P}} \frac{1}{2} \left\|\mathbf{a}\right\|_{2}^{2}$$
$$R_2(D|\theta^{G^\prime}) = \left\| \Sigma' \right\|_{F}$$
$$R_3(D|\theta^{G^\prime}) = \left\| \Gamma \right\|_{F}$$
where $\left\| \cdot \right\|_{2}$ is $L_2$ norm and $\left\| \cdot \right\|_{F}$ is Frobenius norm. $R_1$ regularizes the scale of our embedding by intuition, as well as the scale of the blade and chest jointly, since they are integrated into our embedding. $R_2$ regularizes the scale of the free matrix $\Sigma'$ as well as the scale of the symplectic matrix $\Sigma$, because $\left\| \Sigma \right\|_{F} = \left\| \Sigma' - \Sigma'^T \right\|_{F}$ is upper bounded by 2$\left\| \Sigma \right\|_{F}$. $R_3$ regularizes the scale of the free matrix $\Gamma$, in line with Condition (\ref{eq: norm_condition}) given in Theorem 1.



Therefore, the regularized training objective for a given training dataset is
\begin{equation}
  Q(D,\theta^{G^\prime}) = L(D|\theta^{G^\prime}) - \sum_{i} \lambda_i R_i(\theta^{G^\prime})
\end{equation} 
where $\theta^{G^\prime} := \{\mathbf{a},\mathbf{b},\Sigma',\Gamma\}$ denotes the model parameters and $\lambda$ controls the regularization.

\section{Experiments}

In this section, we first summarize the datasets with a quantitative investigation of the existence of intransitivity. Then, we report the experimental results of our proposed method on several challenging real-world benchmark datasets that consist of pairwise comparisons in social choice and matchups between individual players. 

We used cross validation for parameter tuning in the experiments. Given the dataset in 4-tuple format, we first split the dataset randomly into three folds for cross validation and then identified the unique pairwise interactions and aggregated them. The hyperparameters were the dimensionality of the embedding $d$ and the regularization coefficient $\lambda$. The performance was measured by the average test accuracy $A(\mathbf{D}_{test}|\theta)$, defined by
$$
A(\mathbf{D}_{test}|\theta) = \frac{1}{|\mathbf{D}_{test}|} \sum_{(a,b,n_{a},n_{b})\in\mathbf{D}_{test}} n_{a}\cdot\mathbbm{1}(\hat{o}_a \succ \hat{o}_b) + n_{b}\cdot\mathbbm{1}(\hat{o}_a \prec \hat{o}_b)
$$
where $\mathbbm{1}(\cdot)$ is the indicator function of an event.

We compared our proposed method with three competitive methods, namely the na\"{i}ve method, BT model, and BC model. The \textbf{na\"{i}ve method} estimates the winning probability of each player based on the empirical observations, with $Pr(o_a \succ o_b) = \frac{n_a + 1}{(n_a + 1) + (n_b + 1)}$. If $n_a = n_b$, one player is randomly assigned as the winner. The 
\textbf{BT model} estimates player ability with a scalar representation. The \textbf{BC model} estimates player ability with two multidimensional vectors that are independent of each other.

\subsection{Datasets}
We investigated several challenging benchmark datasets from diversified areas. The datasets are commonly grounded on pairwise comparisons or matchups between objects or players.  SushiA and SushiB \cite{kamishima2003nantonac} are food preference datasets. Jester \cite{globerson2007euclidean} and MovieLens100K \cite{herlocker1999algorithmic} are collective preference datasets in an online recommender system. $\text{ElectionA}_5$ \cite{tideman2006collective} is an election dataset for collective decision making. Within the area of online games, $\text{SF4}_{5000}$ \cite{chen2016modeling} is a dataset collected from professional players and is used to profile the characters in the virtual world. Dota \cite{chen2016modeling} is a dataset of game records produced by a large number of players on an online RPG game platform. 

\subsubsection{Intransitivity in datasets} Quantitative statistics of intransitive relationships in these datasets are presented in Table \ref{summary}. {\it isIntrans} indicates the existence of the intransitivity relationships. {\it Intrans@3} indicates the percentage of intransitive loops that are analogous to the rock-paper-scissors game, where the number of involved players equals 3. In {\it Intrans@3}, the denominator is the total number of directed length-3 loops given by $2\binom{N}{3}$ for a fully observed pairwise dataset. {\it PlayerIntrans@3} is the number of players who are involved in a rock-paper-scissors-like relationship. Both {\it Intrans@3} and {\it PlayerIntrans@3} characterize the intensity of intransitivity, and a higher score indicates more intensive intransitivity in the dataset. In the majority of the seven datasets we investigated, an intransitive relationship exists. Moreover, in five out of the seven datasets, more than half of the players are involved in local intransitive relationships. To this end, we highlight the necessity of modeling the intransitivity, and to the best of our knowledge, this is the first quantitative exploration of the existing intransitive relationships in these prevalent benchmark datasets. 

\begin{table}[h]
\centering
\caption{Summary of real-world datasets}
\label{summary}
\begin{tabular}{|l|c|c|c|c|c|c|}
\hline
DATASET   & No. of Players & No. of Records  & isIntrans & Intrans@3 & No. PlayerIntrans@3    \\ \hline
SushiA    & 10            & 100000                        & x & 0.00\% & 0/10\\ \hline
SushiB    & 100           & 25000                          & \checkmark & 26.87\% & 92/100\\ \hline
Jester    & 100           & 891404                        & \checkmark  & 1.77\%& 97/100 \\ \hline
MovieLens100K & 1682          & 139982                       & \checkmark & 0.19\% & 1130/1682 \\ \hline
$\text{ElectionA}_5$ &16 & 44298 &  \checkmark & 0.44 \%& 6/16\\ \hline 
$\text{SF4}_{5000}$ &35 & 5000 &  \checkmark & 23.86\% & 34/35 \\ \hline 
Dota &757 & 10442 &  \checkmark & 97.58\%& 550/757 \\ \hline 

\end{tabular}
\end{table}

\subsection{Experiments on Real Datasets}
Table \ref{results} shows the experimental results of our proposed method. For all of the four transitivity-rich datasets, SushiB, Jester, $\text{ElectionA}_5$, and $\text{SF4}_{5000}$, we observe improvement in terms of the average test accuracy. In addition to the predictive performance, two practical facts are noteworthy. (a) The observed pairwise interactions in all these datasets are rich, and a $K$-fold cross validation procedure with no data augmentation results in a set of data bins, each of which contains identical players. Therefore, it is guaranteed that the representation of each player in the validation and test bin will be learned by a training set with a size of $K-2$ bins. However, as the number of players grows, the number of records required to accommodate such a cross validation procedure grows quickly. For instance, in the case of the SushiB dataset with 100 players, 25000 pairwise records, Intrans@3 $= 26.87\%$, and PlayerIntrans@3 $= 92/100$, the empirical down sampling for $3$-fold cross validation is sufficient to perform a fully-evidenced prediction of the dominance for all possible player pairs, instead of a random guess caused by the existence of non-observed players in the validation and test bins. (b) Given a sampling scheme that is sufficiently stable to allow the model to give a fully evidenced prediction, a $K$-fold cross validation results in sparser interactions in the bins, which can be indicated by the connectivity of the matchup network, i.e., Borda count or Copeland count for directed graphs. However, in the challenging MovieLens100K and Dota datasets, the resultant heterogeneous interactions between players prevent us from providing evidenced dominance prediction from the observed sparse networks. A trivial solution for such a case is a random guess, which is meaningless for intransitivity recovery. The above two facts hold for all the competitive methods.

\begin{table*}[t]
\centering
\caption{Test accuracy on real-world datasets}
\label{results}
\begin{tabular}{|l|c|c|c|c|}
\hline
DATASET   & Na\"{i}ve             & Bradley-Terry     & Blade-Chest           & Proposed Model                   \\ \hline
SushiA    & $0.6549 \pm 0.0044$ & $0.6549 \pm 0.0021$ & $0.6551 \pm 0.0038$     & $\bm{0.6551 \pm 0.0027}$ \\ \hline
SushiB    & $0.6466 \pm 0.0042$             & $0.6582 \pm 0.0077 $           & $0.6591 \pm 0.0051$   & $\bm{0.6593 \pm 0.0058}$                                           \\ \hline
Jester    & $0.6216 \pm 0.0006$             & $0.6236 \pm 0.0028$        &    $0.6242 \pm 0.0035$                   & $\bm{0.6243 \pm 0.0019}$                             \\ \hline
$\text{ElectionA}_5$ &$0.6507 \pm 0.0031$ & $0.6531 \pm 0.0038$ &  $0.6533 \pm 0.0043$ & $\bm{0.6535 \pm 0.0055}$\\ \hline 
$\text{SF4}_{5000}$ & $0.5297 \pm 0.0102$ & $0.5329 \pm 0.0044 $ &  $0.5329 \pm 0.0062$ & $\bm{0.5355 \pm 0.0080}$ \\ \hline 
\end{tabular}
\end{table*}

\section{Conclusion}
In this paper, we focused on the issue of modeling intransitivity and representation learning for players involved in pairwise interactions. We proposed a generalized embedding formulation for learning the intransitivity-compatible representation from pairwise matchup data, and provided a theoretical characterization of the properties of the proposed formulation in terms of symmetry and expressiveness. We also tailored an efficient solution to the constraint optimization problem and verified the expressiveness of the proposed model by bridging it to former models. A thorough quantitative statistics analysis of the existing intransitivity in various real-world datasets was presented. To the best of our knowledge, it is the first of this kind in the data mining community. The results of the experiments based on real-world datasets show that our method achieves a better performance than the competitive models, i.e., the state-of-the-art BC model.

\bibliography{bibliography}
\bibliographystyle{splncs04}
\clearpage
\end{document}